\documentclass{./templates/svproc}

\usepackage{tikz,xcolor,hyperref}

\definecolor{lime}{HTML}{A6CE39}
\DeclareRobustCommand{\orcidicon}{%
	\begin{tikzpicture}
	\draw[lime, fill=lime] (0,0) 
	circle [radius=0.16] 
	node[white] {{\fontfamily{qag}\selectfont \tiny ID}};
	\draw[white, fill=white] (-0.0625,0.095) 
	circle [radius=0.007];
	\end{tikzpicture}
	\hspace{-2mm}
}

\foreach \x in {A, ..., Z}{%
	\expandafter\xdef\csname orcid\x\endcsname{\noexpand\href{https://orcid.org/\csname orcidauthor\x\endcsname}{\noexpand\orcidicon}}
}

\usepackage{url}
\usepackage{array}
\usepackage{graphics}
\usepackage{tabularx}
\usepackage{graphicx}
\usepackage{hyperref}
\usepackage{amsmath}
\usepackage{algorithm}
\usepackage[noend]{algpseudocode}

\begin{document}

\mainmatter              
\title{LaDiff ULMFiT: A Layer Differentiated training approach for ULMFiT}
\titlerunning{LaDiff ULMFiT}  
%
\author{Mohammed Azhan \orcidA{}\inst{1} \and Mohammad Ahmad \orcidB{}\inst{2}}
\authorrunning{Mohammed Azhan et al.} 
%
\tocauthor{Mohammed Azhan, Mohammad Ahmad}
\institute{Dept. of Electrical Engineering, Jamia Millia Islamia, New Delhi, India,\\
\email{mohd178974@st.jmi.ac.in},\\ WWW home page:
\href{https://azhanmohammed.netlify.app/}{https://azhanmohammed.netlify.app/}
\and
Dept. of Electronics and Communication Engineering, Jamia Millia Islamia, New Delhi, India,\\
\email{mohammad178576@st.jmi.ac.in}, \\WWW home page:
\href{https://mohammadahmad242.netlify.app/}{ahmadkhan242.github.io}}

\maketitle              

\begin{abstract}
In our paper we present Deep Learning models with a layer differentiated training method which were used for the SHARED TASK @ CONSTRAINT 2021 sub-tasks COVID19 Fake News Detection in English and Hostile Post Detection in Hindi. We propose a Layer Differentiated training procedure for training a pre-trained ULMFiT\cite{howard2018universal} model. We used special tokens to annotate specific parts of the tweets to improve language understanding and gain insights on the model making the tweets more interpretable. The other two submissions included a modified RoBERTa model and a simple Random Forest Classifier. The proposed approach scored a precision and f1-score of 0.96728972 and 0.967324832 respectively for sub-task COVID19 Fake News Detection in English. Also, Coarse Grained Hostility f1 Score and Weighted Fine Grained f1 score of 0.908648 and 0.533907 respectively for sub-task Hostile Post Detection in Hindi. The proposed approach ranked 61st out of 164 in the sub-task "COVID19 Fake News Detection in English" and 18th out of 45 in the sub-task "Hostile Post Detection in Hindi". The complete code implementation can be found at: \href{https://github.com/sheikhazhanmohammed/AAAI-Constraint-Shared-Tasks-2021}{GitHub Repository}\footnote{\href{https://github.com/sheikhazhanmohammed/AAAI-Constraint-Shared-Tasks-2021}{https://github.com/sheikhazhanmohammed/AAAI-Constraint-Shared-Tasks-2021}}
\end{abstract}
\section{Introduction}
COVID-19 was declared as a global health pandemic by the WHO, and it can be very well noticed that social media has played a very significant role much before the spread of the virus. As various countries around the world went into lockdown for long periods, it was noticed that social media became a very important platform for people to share information, post their views and emotions in short amount of texts. It has been seen that a study of these texts have resulted in various novel applications which are not only limited to, political opinion detection as seen in \cite{Maynard2012}, stock market monitoring as seen in \cite{Abdullah2013}, and analysing user reviews of a product as seen in \cite{DBLP:journals/corr/abs-1904-04096}. The wide usage of figurative language like hashtags, emotes, abbreviations, and slangs makes it even more difficult to comprehend the text being used on these social platforms, making Natural Language Processing a more challenging task.
It has been seen that techniques like Latent Topic Clustering \cite{Lee2018}, Cultivating deep decision trees \cite{Ignatov2017}, performing Fine grained sentiment analysis \cite{DBLP:journals/corr/abs-1904-04096}, and ensemble techniques \cite{Balikas2017} have given competitive results in language understanding tasks in NLP. In this paper we present a similar Deep Learning technique which competed in AAAI SHARED TASK @ CONSTRAINT 2021 'COVID 19 Fake News Detection in English' and 'Hostile Post Detection in Hindi'. The overview of above Shared Task has been explain in this\cite{patwa2021overview}. We explored differentiated layer training technique, where different sections of the layers were frozen and unfrozen during the training. This was combined with the training procedure as discussed in ULMFiT \cite{howard2018universal}. The complete training procedure is explained in the coming sections. The paper is divided into sections, the next section discusses the task at hand, details of the dataset provided and the preprocessing steps that were taken.

\section{Overview}
This section contains details of the given task, the dataset provided, and the preprocessing steps taken to clean the dataset.
\subsection{Task Description and Dataset}
\textbf{Task Definition Sub-task 1}  This subtask focuses on the detection of COVID19-related fake news in English. The sources of data are various social-media platforms such as Twitter, Facebook, Instagram, etc. Given a social media post, the objective of the shared task is to classify it into either fake or real news. 
The dataset provided for the task is discussed in \cite{patwa2020fighting}. The dataset contains a total of 6420 labeled tweets for training, 2140 labeled tweets for validation and 2140 unlabeled tweets were given during the test phase. The complete class distribution for the dataset is shown in Fig. 1 (a).
The image shows that the distribution of the classes was almost balanced, hence no under-sampling or over-sampling techniques were used during the preprocessing to balance the dataset.\\
\textbf{Task Definition Sub-task 2} This subtask focuses on a variety of hostile posts in Hindi Devanagari script collected from Twitter and Facebook. The set of valid categories are fake news, hate speech, offensive, defamation, and non-hostile posts. It is a multi-label multi-class classification problem where each post can belong to one or more of these hostile classes.
The dataset for this  sub-task covers four hostility dimentions: fake news, hate speech, offensive, and defamation posts, along with a non-hostile label. Dataset is multi labelled due to overlap of different hostility classes. The dataset is further described here \cite{bhardwaj2020hostility} .
The dataset provided 5728 labeled posts for training, 811 labeled post for validation, and 1653 unlabeled for test phase. The labeled distribution for train set is shown in Fig. 1 (b).
\begin{figure}[h]
\centering
\includegraphics[scale=0.35]{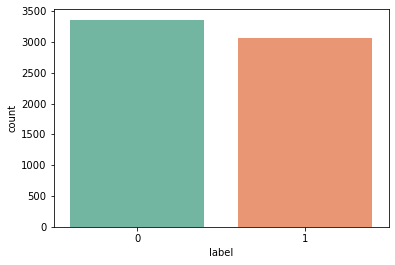}
\includegraphics[scale=0.35]{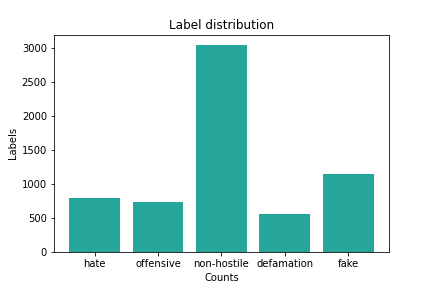}
\caption{Label distribution for training dataset (a)"COVID19 Fake News Detection" (b)"Hostile Post Detection in Hindi"}
\end{figure}
\subsection{Preprocesing}
The various steps used during the preprocessing of the dataset are mentioned below.
\subsubsection{Replacing Emojis}  Since tweets from twitter are mostly accompanied with graphics (emojis) which are supposed to help a user express his thoughts, our first task was to replace these emojis with their text counterpart. While a machine cannot understand the emoji, it's text counterpart can easily be interpreted as discussed in \cite{2016} and \cite{Azhan2020}. We used the emoji library\footnote{https://pypi.org/project/emoji/} for converting emojis to their English textual meanings. For the Hindi dataset we created our own library 'Emot Hindi' \footnote{https://github.com/ahmadkhan242/emot\_hindi} similar to the emoji library discussed above which contains emojis and their Hindi textual meanings. This was a common step for both sub-tasks. A few examples of sample emojis and their meanings are shown in Fig. 2.
\begin{figure}[h]
\begin{center}
\includegraphics[]{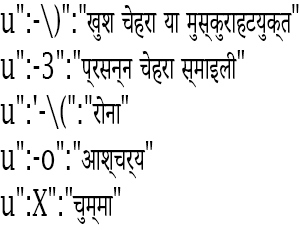}
\includegraphics[]{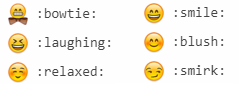}
\caption{Example: Emoji and text counterpart (a)Emoji to Hindi (b) Emoji to English}
\end{center}
\end{figure}
\subsubsection{Addressing hashtags} Hashtags are word or phases preceded by a hash sign '\#' which are used to identify texts regarding a specific topic of discussion. It has been seen that the attached hashtags to a post or tweet tell what the text is relevant to, this has been discussed in \cite{Azhan2020} and \cite{Alfina2017}. For the given tweets a white space was added between the hash symbol and the following word for the model to comprehend it easily. This was also a common step for both sub-tasks.
 \subsubsection{Adding special tokens} We replaced specific parts of the text with special tokens as discussed in the fastai library\footnote{https://docs.fast.ai/text.core.html}. The special tokens and their usage are mentioned in the list below.
\begin{itemize}
    \item \textbf\{TK\_REP\} This token was used to replace characters that were occurring more than thrice repeatedly. This special token was used for both sub-tasks. For example 'This was a verrrryyyyyyy tiring trip' will be replaced with 'This was a ve\{TK\_WREP\} 4 r \{TK\_WREP\} 7 y tiring trip'.
    \item \textbf{\{TK\_WREP\}} This token was used to replace words occuring three or more times consecutively. This special token was used for both sub-tasks. For example 'This is a very very very very very sad news' will be replaced with 'This is a \{TK\_WRPEP\} 5 very sad news'.
    \item \textbf{\{TK\_UP\}} This token was used to replace words using all caps. Since the Devnagri script used for Hindi has no uppercase alphabetsm this special token was used for the English sub-task only. For example 'I AM SHOUTING' becomes '\{TK\_UP\} i \{TK\_UP\} am \{TK\_UP\} shouting'.
    \item \textbf{\{TK\_MAJ\}} Used to replace characters in words which started with an upper case except for when it is the starting of a sentence. Again, this special token was used for the English subtask only. For example, 'I am Kaleen Bhaiya' becomes 'i am \{TK\_MAJ\} kaleen \{TK\_MAJ\} bhaiya'.
\end{itemize}
\subsubsection{Normalization} These steps included removing extra spacing between words, correcting hmtl format from texts if any, adding white space between special characters and alphabets, and replacing texts with lower case. The above preprocessing steps were taken for both subtasks.
\subsubsection{Tokenization} Once the preprocessing of the dataset was complete, we performed tokenization. For the ULMFiT training the ULMFiT tokenizer was used, similarly the text for the customized RoBERTa model was tokenized using RoBERTa tokenizer, and for the Random Forest Classifier (English and Hindi sub-task) and Linear Regression (Hindi sub-task) the text was tokenized using the nltk library for both the languages.

\section{Model Description}
Next, we provide an in detail description of the training strategies that were used to achieve the results. The test results obtained using each technique is mentioned in the results section. Each technique is discussed in the coming sub-sections.
\subsection{Layer Differentiated ULMFiT Training}
As discussed in \cite{howard2018universal} inductive training has shown incredible performance in Computer Vision tasks where the model is first pretrained on large datasets like ImageNet, MS-COCO, and others. The same idea was implemented during the training of the ULMFiT model, only it was modified using a pretrained language model. Traditional transfer learning language models used to pretrain the language model on a relatively larger dataset, this language model was then used to create the classifier model which will again pretrain on the large dataset, at the final step the classifier model was fine-tuned on the target dataset. ULMFiT introduced LM Pretraining and Fine-tuning to make sure that the language model used to pretrain the classifier consisted of extracted features from the target domain. This part of the training procedure is exactly same as discussed in \cite{howard2018universal}. The image below shows the training of both Language model and classifier as in \cite{howard2018universal}.
\begin{figure}[ht]
\centering
\includegraphics[scale=0.22]{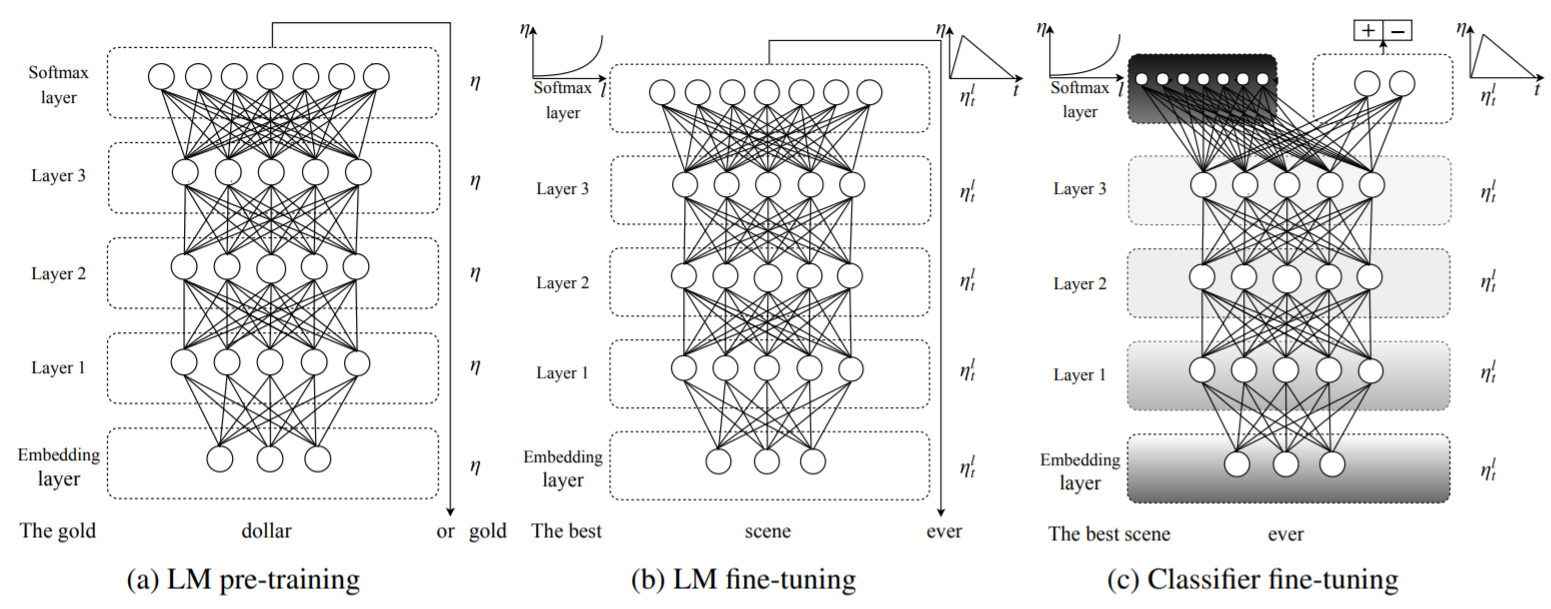}
\caption{ULMFiT Traditional Training}
\end{figure}
We introduced a layer differentiated training procedure, which gradually unfreezed the layers for training them. This differentiated training procedure was implemented for training both, the language model and the classifier model for both of the sub-tasks.
Fig. 4 shows a plot between the training and validation losses as the training progressed for the English sub-task. The graph shows a spike after every 100 batches which is then followed by a sharp decline. These spikes are the parts where the layers were unfreezed. As the layers were unfreezed, the untrained layers led to an increase in the training loss, which gradually decreased as the training progressed. This also made sure that the final layers were trained longer as compared to initial layers so that the initial layers don\'t start overfitting and the model doesn\'t drops out any important features.   
\begin{figure}[h]
\centering
\includegraphics[scale=0.75]{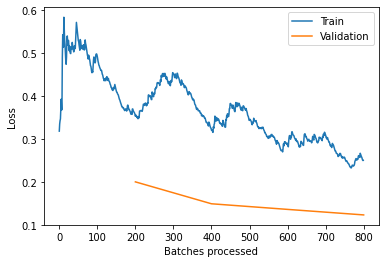}
\caption{Loss vs Batches Progressed: LaDiff ULMFiT}
\end{figure}
This concludes our discussion for the LaDiff ULMFiT training. We now move forward with our next technique.
\subsection{Customized RoBERTa}
RoBERTa\cite{DBLP:journals/corr/abs-1907-11692} is a robustly optimized pretraining approach designed for BERT\cite{DBLP:journals/corr/abs-1810-04805}. BERT stands for Bidirectional Encoder Representations from Transformers, and it introduced the use of transformers for language training tasks. RoBERTa aimed at improvising the training methodology as introduced in BERT using dynamic masking, provising full sentences rather than using next sentence prediction, training with a large number of batches having small sizes and a larger byte-level Byte-Pair Encoding. For our customized model, we used the RoBERTa uncased model pre-trained on various larger twitter dataset. We then added a few customized layers to the model. This training procedure was implemented on the English sub-task only. 
\subsection{Random Forest Classifiers and Logistic Regression}
While the above two approaches have shown how language modelling and using text transformers give exceptionally high performance, our idea behind trying these approach was to understand where do simple language classifiers lack as compared to deep neural networks. While the baseline results as presented in the English dataset paper \cite{patwa2020fighting} and Hindi dataset paper \cite{bhardwaj2020hostility} use an SVM Classifier, we decided to use various Machine Learning techniques, and submit the one which has the highest score in the validation set. In our case, we achieved the best results using a Random Forest Classifier, having n\_estimators set as 1000, min\_samples\_split as 15 and a random\_state of 42. The same classifier hyper-parameters were passed to both of the classification models and trained separately.
The Logistic Regression Classifier was used only for the Hindi sub-task.
This brings an end to our discussion for the various approaches used. We now move forward to the results obtained and compare them with the available baseline results\cite{patwa2020fighting} \cite{bhardwaj2020hostility}.

\section{Results}
We first present the results obtained for the English sub-task "COVID19  Fake  News  Detection  in English".The table given below gives the accuracy, precision, recall and f1-score of our approaches and compares them with the available baseline results.
\begin{table}[]
\begin{tabular}{|l|r|r|r|r|}
\hline
\textbf{Model} & \multicolumn{1}{l|}{\textbf{Accuracy}} & \multicolumn{1}{l|}{\textbf{Precision}} & \multicolumn{1}{l|}{\textbf{Recall}} & \multicolumn{1}{l|}{\textbf{f1-score}} \\ \hline
\textbf{LaDiff ULMFiT} & \textbf{0.96728972} & \textbf{0.967908486} & \textbf{0.96728972} & \textbf{0.967324832} \\ \hline
\textbf{Baseline Model} & 93.46 & 93.46 & 93.46 & 93.46 \\ \hline
\textbf{Customized RoBERTa} & 0.929906542 & 0.929906542 & 0.929906542 & 0.929906542 \\ \hline
\textbf{Random Forest Classifier} & 0.91728972 & 0.917382499 & 0.91728972 & 0.917311831 \\ \hline
\end{tabular}
\caption{Comparison Results on Test Set: LaDiff ULMFiT vs Customized RoBERTa vs Random Forest Classifier vs Baseline Model- Sub-task 1}
\end{table}
Our best approach, LaDiff-ULMFiT ranked 61st out of 167 submissions on the final leaderboard.\\
We now present our results for the Hindi sub-task "Hostile Post Detection in Hindi" shown in Table 2.
\begin{table}[]

\begin{tabular}{|m{10em}|m{4.5em}|m{6em}|m{3.5em}|m{3.5em}|m{5em}|m{5em}|}
\hline
\textbf{Model} &
  \textbf{Coarse Grained Hostility f1 Score} &
  \textbf{Defamation f1 Score} &
  \textbf{Fake f1 Score} &
  \textbf{Hate f1 Score} &
  \textbf{Offensive f1 Score} &
  \textbf{Weighted Fine Grained f1 Score} \\ \hline
\textbf{LaDiff ULMFiT}            & \textbf{90.87} & 27.31        & \textbf{73.83} & 44.93       & \textbf{51.39} & \textbf{0.53} \\ \hline
\textbf{Baseline Results}         & 84.11             & \textbf{43.57} & 68.15             & \textbf{47.49} & 41.98             &                   \\ \hline
\textbf{Logistic Regression}      & 76.56          & 24.8          & 54.71          & 40.65       & 40.58          & 42.74          \\ \hline
\textbf{Random Forest Classifier} & 76.56          & 24.8          & 54.71          & 40.65       & 40.58          & 42.74          \\ \hline
\end{tabular}

\caption{Comparison Results on Test Set: LaDiff ULMFiT vs Logistic Regression vs Random Forest Classifier vs Baseline Results- Sub-task 2}
\label{tab:my-table}
\end{table}
The results ranked 18th for the Coarse Grained f1 Score and 25th for the Fine Grained f1 Score. We now proceed with our conclusions.
\section{Conclusions}
From the achieved results as shown in Table 2, the following conclusions can be drawn:
\begin{itemize}
    \item Fine-tuned language model used with a simple classifier (LaDiff-ULMFiT) outperforms transformers used with sophisticated networks (Customized RoBERTa).
    \item The losses trend seen in Fig. 4 also signifies the fact that target domain fine tuned on a pre-trained model done at when trained at gradual steps leads to faster decrease in losses.
    \item We also conclude that, tweets containing hashtags and short texts can also be confidently classified using Machine Learning techniques.
\end{itemize}
Finally, we make all our approaches and their source codes completely available for the open source community, to reproduce the results and facilitate further experimentation in the field.

%
%
\bibliographystyle{plain}
\bibliography{bibliography}

\end{document}